\documentclass[sigconf,nonacm]{acmart}
\usepackage{amsmath}
\usepackage{colortbl}
\definecolor{aegishl}{HTML}{E9F1FB}

\title{Keep Compliance in a Pseudonymous World: A Federated Mixture of Experts for Illicit-Address Screening}
\author{Qian'ang Mao}
\affiliation{%
  \institution{Nanjing University}
  \city{Nanjing}
  \country{China}
}
\author{Shujing Qiu}
\affiliation{%
  \institution{Nanjing University}
  \city{Nanjing}
  \country{China}
}
\author{Yuxuan Zhang}
\affiliation{%
  \institution{Nanjing University}
  \city{Nanjing}
  \country{China}
}
\author{Jiaqi Yan}
\authornote{Corresponding author.}
\affiliation{%
  \institution{Nanjing University}
  \city{Nanjing}
  \country{China}
}

\begin{document}
\begin{abstract}
Public blockchains expose transaction graphs but hide real-world identity, leaving Virtual Asset Service Providers (VASPs) with fragmented views of anti-money-laundering (AML) risk. Standard federated aggregation treats participants as globally interchangeable even though their expertise is address- and typology-dependent. We introduce \textbf{AEGIS}, an auditable federated mixture of experts for illicit-address screening. AEGIS transfers a FedSage checkpoint into a Dual-Path GraphSAGE base and combines seven laundering-motif features with one residual expert per client. Dual-Track Expert Adaptation (DTEA) selects a client-locally supervised track only when it clears a fixed validation margin, while a multi-anchor bank selects its mixture on validation data; both decisions are frozen before scoring. We benchmark AEGIS against federated learners applicable when raw transaction data cannot leave a VASP, including the 2024--2026 FedTAD, FedTGP, FedAux, and FedGraph-VASP frameworks, on seven weakly supervised sanctions graphs and a public XBlock-derived Ethereum phishing benchmark. Across eight scenarios, AEGIS improves mean F1 from 0.751 for the strongest complete baseline, FedGraph-VASP, to 0.839; it leads seven scenarios and ties the eighth.
\end{abstract}


\maketitle

\section{Introduction}

Public blockchains reveal transaction histories but not the entities behind cryptographic addresses \cite{bonneau2015BitcoinCryptocurrencies,tschorsch2016BitcoinBeyond}. Illicit flows cross exchanges, bridges, and wallets, while each VASP sees only its own customers and local labels. The setting is a serious cross-domain security problem: the ledger is public and pseudonymous, the labels remain inside each institution, and illicit paths run across that boundary. Pooling those private records is often infeasible, so institutions have to collaborate without exchanging raw customer data. Federated learning addresses such silos by exchanging parameters or predictions instead of raw records.

Consider a queried address that receives funds along several laundering-motif routes, including a peel chain, a mixer star, and a bridge fan-out, while each client's labels remain local. This yields our central research question: \emph{how can VASPs screen illicit addresses on this public graph without pooling client labels?}

To address this question, we propose \textbf{AEGIS} (\emph{Anchored Expert Gating for Illicit-address Screening}), a federated Mixture-of-Experts (MoE). A shared GraphSAGE base is initialized from FedSage; each VASP supplies a residual expert; and an attention router selects a sparse query-dependent subset and assigns query-dependent normalized weights. Seven explicit AML motifs add inspectable domain priors. DTEA trains standard and client-locally supervised expert tracks, selecting one track on validation data before inference. A validation-selected anchor bank chooses one convex mixture of the core and two shared federated anchors and freezes it for the scoring window.

Our contributions are: (1) a federated mixture of residual experts on a transferred GraphSAGE base, with client labels kept local; removing those residuals lowers macro F1 by 0.010; (2) DTEA, a validation-gated switch that deploys the client-local track only when it clears a fixed margin; removing DTEA lowers macro F1 by 0.006; (3) an auditable validation-selected multi-anchor protocol with explicit motifs and a prespecified low-positive guard; and (4) an evaluation against eight federated baselines on eight blockchain scenarios with a strict single-component ablation. Extended definitions, settings, and per-scenario results are in the Appendix.

\section{Related Work}

Blockchain AML methods exploit graph topology and transaction behavior to cluster addresses and detect suspicious activity \cite{ermilov2017AutomaticBitcoinClustering,weber2019AntiMoneyBitcoin,wang2024UnconfirmedTransactions,zhou2024VisualAnalysisMoney}. Most assume centralized graph access, whereas consortium screening combines public topology with private, institution-specific context.

\paragraph{Cross-VASP AML.}
Application-specific collaboration has progressed from proof of concept to privacy-protected deployment. \citet{suzumura2022federatedfinancialcrime} formulate financial-crime detection as federated graph learning across institutions and report a gain over isolated models on the UK FCA TechSprint data. \citet{kanamori2022japanesebanks} demonstrate privacy-preserving federated training on real transactions from five Japanese banks, showing that cross-bank learning can surface fraud missed locally. More recently, \citet{nguyen2025privacyaml} combine differential privacy and secure model aggregation for AML, but represent each institution through a tabular random-forest learner rather than a shared transaction graph. Most directly, the concurrent FedGraph-VASP preprint exchanges encrypted boundary-node embeddings between VASP subgraphs and finds that its advantage depends strongly on cross-silo connectivity \cite{commey2026fedgraphvasp}. These studies establish both the value and the privacy cost of cross-institution collaboration. They nevertheless use a global model or fixed boundary exchange. AEGIS keeps client labels local and freezes the validation-selected track, anchor mixture, guard, and threshold before scoring.

Federated learning exchanges parameters or predictions rather than raw records. FedAvg performs data-weighted aggregation \cite{mcmahan2017communication}; FedProx regularizes local drift \cite{li2020fedprox}; and FedSage adapts GraphSAGE \cite{hamilton2017inductive} to federated subgraphs \cite{zhang2021subgraph}. Recent graph-federated methods use topology-aware averaging (FedGTA) \cite{li2023fedgta}, data-free distillation (FedTAD) \cite{zhu2024fedtad}, or auxiliary projections (FedAux) \cite{zhuo2025fedaux}; the general FedTGP method learns trainable global prototypes \cite{zhang2024fedtgp}. None combines a transferred GraphSAGE base with explicit laundering motifs and a validation-frozen anchor mixture. Robust aggregation addresses poisoned or poor contributors \cite{yin2018byzantine}, but does not model complementary client expertise.

MoE models learn specialized experts and a gate that selects them per input \cite{jacobs1991adaptive,shazeer2017outrageously}. Sparse routing reduces computation while load-balancing discourages expert collapse \cite{fedus2022switch}. AEGIS uses one residual expert per client on a shared GraphSAGE base, with client-local expert supervision, a validation-gated standard track, and transferred federated anchors.

\section{Problem and Threat Model}
\label{sec:problem}

Let $G=(V,E)$ be a directed transaction graph and $V_C=\{1,\ldots,n\}$ a VASP consortium. Client $i$ observes a local subgraph $G_i$, public on-chain features, and private labels. Given address $x$, the binary task predicts sanctioned/phishing or weakly labeled laundering-intermediary risk versus normal activity. Raw records, labels, and local membership are not shared.

The scorer combines shared predictors with one residual expert per client. The design has two practical objectives: \emph{locality}, because client labels cannot be pooled; and \emph{auditability}, because the track, anchor mixture, guard, and threshold must be fixed from training and validation evidence before scoring.

The coordinator is honest-but-curious: it follows the protocol but may infer whether a queried address belongs to a VASP's data. AEGIS keeps raw records and local labels at clients.

\section{AEGIS}
\label{sec:design}

\begin{figure*}[t]
  \centering
  \includegraphics[width=\textwidth]{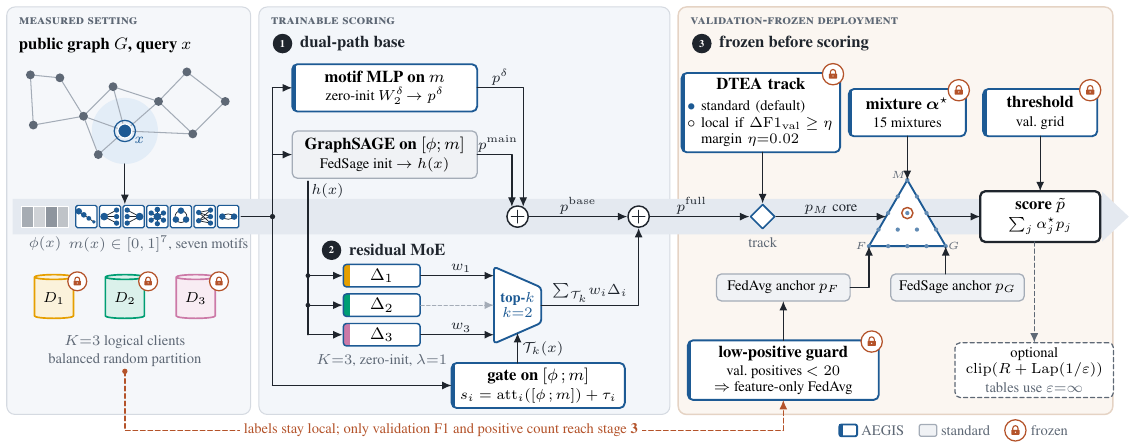}
  \Description{Architecture of AEGIS: public features and motifs feed a dual-path GraphSAGE base and sparse residual experts, followed by a frozen validation-selected deployment record.}
  \caption{AEGIS overview. \textbf{Left (measured setting):} public graph $G$ and query $x$ yield behavioral features $\phi(x)$ and seven motif scores $m(x)\in[0,1]^7$; $K=3$ clients hold labels locally (locks). \textbf{Center (trainable scoring):} (1)~dual-path base forms $p^{\mathrm{base}}=p^{\mathrm{main}}+p^{\delta}$ from GraphSAGE (FedSage init) and a zero-init motif MLP; (2)~residual MoE experts $\Delta_i$ are gated on $[\phi;m]$, and top-$k{=}2$ adds weighted residuals to get $p^{\mathrm{full}}$. \textbf{Right (frozen before scoring):} validation F1 and the validation positive count lock DTEA track choice (local only if $\Delta\mathrm{F1}_{\mathrm{val}}\ge\eta{=}0.02$), the 15-point $\boldsymbol\alpha^\star$ mixture of $p_M$ / FedAvg / FedSage anchors, the low-positive guard, and the decision threshold. Blue outlines mark AEGIS-specific components; gray marks standard ones; red locks mark decisions frozen at deployment.}
  \label{fig:framework}
\end{figure*}

\subsection{Auditable Motif Priors and Dual-Path Base}

Behavioral statistics capture transaction volume, degree, and timing but not recognizable laundering patterns. We therefore compute seven bounded motif scores: peel chain, fan-out, fan-in, mixer star, wash cycle, bipartite hub, and ping-pong. Each score lies in $[0,1]$ and reads only the observed ego graph. Peel chain is the fraction of temporally adjacent in/out pairs, within one hour, whose value ratio lies in $[0.80,0.99]$. Fan-out combines out-neighbor diversity with the fraction of outputs below 20\% of the representative input, and fan-in is the symmetric consolidation score. Mixer star decays with the coefficient of variation of neighbor value and saturates at 50 neighbors. Wash cycle is the fraction of out-neighbors on a length-three return path. Bipartite hub multiplies in/out value balance by log-scaled throughput. Ping-pong is the fraction of counterparties present in both neighbor sets.

Let $\phi(x)\in\mathbb{R}^{d}$ be behavioral features and $m(x)\in[0,1]^7$ the motif vector. The shared base has a two-layer GraphSAGE main path and a motif correction:
\begin{align}
p^{\mathrm{main}}(x)&=\operatorname{SAGE}^{(2)}([\phi(x);m(x)],G),\\
p^{\delta}(x)&=W_2^{\delta}\operatorname{ReLU}(W_1^{\delta}m(x)),\quad W_2^{\delta}|_{t=0}=0,\\
p^{\mathrm{base}}(x)&=p^{\mathrm{main}}(x)+p^{\delta}(x).
\end{align}
The zero initialization makes the motif path an additive correction and provides an exact transferred-anchor checkpoint before fine-tuning.

\subsection{Anchored Training and Dual-Track Adaptation}
\label{sec:fedmoe}

\textbf{Stage 1} trains FedSage and transfers its two GraphSAGE layers to $p^{\mathrm{main}}$; the zeroed motif path makes the initial base prediction identical to that internal checkpoint. A validation check disables the motif path when it hurts the main path. \textbf{Stage 2} attaches one zero-initialized residual head $\Delta_i$ per VASP and jointly fine-tunes the base, experts, and router. The full logits are
\begin{equation}
p^{\mathrm{full}}(x)=p^{\mathrm{base}}(x)+\sum_{i\in\mathcal{T}_k(x)}w_i(x)\Delta_i(h(x)),
\label{eq:full}
\end{equation}
where $h(x)$ is the shared embedding.

For learned expert key $e_i$, sparse routing uses
\begin{align}
s_i(x)&=\frac{(W_Q[\phi(x);m(x)])^\top e_i}{\sqrt{d_g}}+\tau_i,\\
w_i(x)&=\frac{\exp s_i(x)}{\sum_{j\in\mathcal{T}_k(x)}\exp s_j(x)}\mathbf{1}[i\in\mathcal{T}_k(x)],
\end{align}
where $\tau_i$ is a learned expert prior and $\mathcal{T}_k(x)$ contains the top-$k$ scores. Inspired by load balancing in sparse MoE routing \cite{fedus2022switch}, the standard track minimizes class-weighted cross-entropy plus the quadratic balancing penalty $\mathcal L_{\mathrm{std}}=\operatorname{CE}(p^{\mathrm{full}},y)+\beta\sum_i K\bar w_i^2$, with $\beta=.01$. To prevent a useful client expert from being starved by early routing, DTEA trains a second track with
\begin{align}
\mathcal L_{\mathrm{local}}&=\mathcal L_{\mathrm{std}}+\frac{\gamma}{K}\sum_{i=1}^{K}\mathcal L_i^{\mathrm{exp}},\\
\mathcal L_i^{\mathrm{exp}}&=\operatorname{CE}\!\left(\operatorname{sg}[p^{\mathrm{base}}(D_i)]+\Delta_i(h(D_i)),y_i\right),
\end{align}
where $D_i$ remains local, $\operatorname{sg}$ blocks the direct base-logit gradient while the residual branch adapts the shared embedding and expert $i$, and $\gamma=.25$. Both tracks use the same anchor-bank procedure below. DTEA deploys the local track only when its post-anchor validation F1 exceeds the standard track by the fixed margin $\eta=.02$; otherwise it falls back to the standard track. This choice is made once before scoring, not per query, and only the selected track and its top-$k$ experts are used at inference.

\subsection{Validation-Selected Multi-Anchor Bank}
\label{sec:anchor-bank}

The core MoE is complementary to simpler federated predictors: a feature-only model can be more stable when positives are extremely scarce, while graph propagation can help in denser regimes. We therefore train two additional shared anchors, feature-only FedAvg and graph-aware FedSage, on independent copies of the same training data. For core, feature, and graph probabilities $p_M,p_F,p_G$, respectively, the final predictor is
\begin{equation}
\widetilde p_{\boldsymbol\alpha^\star}(x)=\alpha^\star_Mp_M(x)+\alpha^\star_Fp_F(x)+\alpha^\star_Gp_G(x),
\end{equation}
where validation selects one deployment-level vector $\boldsymbol\alpha^\star$ from the fixed simplex grid $\alpha_j\in\{0,.25,.5,.75,1\}$, $\sum_j\alpha_j=1$. Among the 15 candidate mixtures, validation F1 is primary, PR-AUC breaks ties, and remaining ties prefer larger $\alpha_M$. Unlike $w_i(x)$, $\boldsymbol\alpha^\star$ is query-invariant and remains frozen throughout the scoring window. The decision threshold is selected on validation data by the common fixed grid. When validation contains fewer than 20 positives, searching 15 mixtures is high variance; the prespecified guard activates the feature-only anchor and selects only its threshold. The two anchors add training cost and two forward passes.

\paragraph{Validation-only deployment procedure.}
The procedure is prespecified. First, the standard and client-local tracks are trained on identical splits, while FedAvg and FedSage anchors are trained on independent copies of the same training data. Second, each track enumerates the anchor mixtures and common threshold grid on validation data; the low-positive guard is applied before mixture search. Third, DTEA selects the local track only if its post-anchor validation F1 exceeds the standard track by $\eta$. The chosen track, $\boldsymbol\alpha^\star$, guard state, and threshold are then frozen for the scoring window, while $w_i(x)$ continues to vary across queries. This ordering prevents scoring-period outcomes from deciding whether a track, anchor, or fallback is active.

\begin{table}[t]
\centering
\small
\setlength{\tabcolsep}{3pt}
\caption{Evaluation graphs. Pos. is the supervised positive class; most XBlock neighbor nodes are unlabeled.}
\label{tab:data}
\begin{tabular}{llrrr}
\toprule
\textbf{Scenario} & \textbf{Chain} & \textbf{Nodes} & \textbf{Edges} & \textbf{Pos.}\\
\midrule
Iran-T & TRON & 9,284 & 23,431 & 483\\
Russia-T & TRON & 12,265 & 37,351 & 830\\
Terror-T & TRON & 107,390 & 357,500 & 2,660\\
Israel-T & TRON & 672,466 & 2,909,493 & 24,769\\
Iran-E & Ethereum & 234,768 & 413,320 & 198\\
Russia-E & Ethereum & 51,093 & 152,213 & 89\\
Terror-E & Ethereum & 515,593 & 1,043,132 & 198\\
XBlock & Ethereum & 520,938 & 659,670 & 4,321\\
\bottomrule
\end{tabular}
\end{table}

\begin{table*}[t]
\centering
\small
\setlength{\tabcolsep}{3.5pt}
\caption{Positive-class test F1 on eight scenarios, averaged over repeated runs. Bold is best per scenario (including ties) and the shaded row is AEGIS.}
\label{tab:t1_baselines}
\begin{tabular}{lrrrrrrrrr}
\toprule
\textbf{Method} & \textbf{Iran-T} & \textbf{Russia-T} & \textbf{Terror-T} & \textbf{Israel-T} & \textbf{Iran-E} & \textbf{Russia-E} & \textbf{Terror-E} & \textbf{XBlock} & \textbf{Mean} \\
\midrule
FedAvg & .5966 & .6312 & .7531 & .8121 & .7349 & .6038 & .6886 & .9080 & .7160 \\
FedProx & .5788 & .5727 & .8078 & .8218 & .7843 & .4710 & .7880 & .8873 & .7140 \\
FedSage & .5788 & .5727 & .8082 & .8220 & .7843 & .4710 & .7880 & .8873 & .7140 \\
FedGTA & .6829 & .7821 & .8051 & .8163 & .6997 & .4435 & .7902 & .8904 & .7388 \\
FedTAD & .5492 & .5118 & .7547 & .8038 & .6782 & .3972 & .7455 & .8872 & .6660 \\
FedTGP & .5514 & .4746 & .7205 & .7904 & .7791 & .5574 & .7993 & .8599 & .6916 \\
FedAux & .5967 & .4933 & .8196 & .8232 & .8089 & \textbf{.6412} & .7748 & .8931 & .7314 \\
FedGraph-VASP & .8031 & .8768 & .8386 & .8099 & .6945 & .3427 & .7540 & .8851 & .7506 \\
\rowcolor{aegishl}\textbf{AEGIS} & \textbf{.8727} & \textbf{.9179} & \textbf{.8967} & \textbf{.8390} & \textbf{.8117} & .6038 & \textbf{.8370} & \textbf{.9320} & \textbf{.8389} \\
\bottomrule
\end{tabular}
\end{table*}

\begin{figure*}[t]
  \centering
  \includegraphics[width=\textwidth]{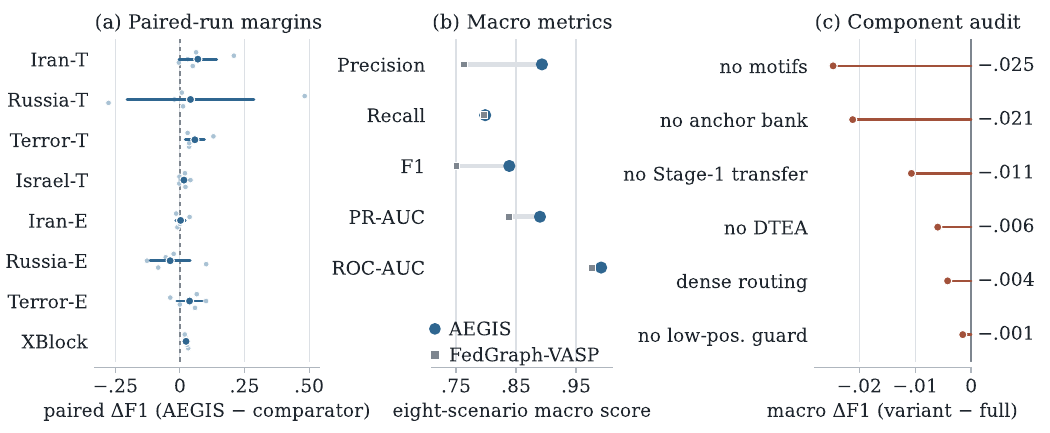}
  \Description{Paired run-level F1 margins per scenario, macro metrics for AEGIS and FedGraph-VASP, and macro-F1 changes of strict ablations.}
  \caption{Result diagnostics from repeated runs. (a) Each row compares AEGIS with the federated method having the highest mean F1 for that scenario; pale points are paired runs and the dark point with its range bar is the mean $\pm$ standard deviation. (b) Eight-scenario macro metrics for AEGIS and the strongest comparator. (c) Macro-F1 changes in the strict single-component ablation, ordered by magnitude; every change is negative.}
  \label{fig:result-diagnostics}
\end{figure*}

\section{Experimental Setup}
\label{sec:experiments}

We organize the evaluation around four questions: whether AEGIS improves over federated alternatives across heterogeneous graphs (Q1), whether gains hold across runs and threshold-independent metrics (Q2), whether each component contributes under a strict single-component ablation (Q3), and which choices validation selects for deployment, together with their systems cost (Q4).

\textbf{Data.} Seven graphs combine public OFAC digital-currency designations \cite{ofacSanctionsSearch} and NBCTF records \cite{nbctfCrypto} with public Ethereum and TRON USDT neighborhoods. They span 9,284--672,466 nodes and 89--24,769 weak positive labels. A listed seed is sanctioned and an unlabeled one-hop neighbor is a laundering intermediary; both are positive, reflecting that direct exposure to a sanctioned address is the primary risk signal in sanctions screening \cite{chainalysis2020exposure,elliptic2023hops}. Structurally normal addresses are negative, and high-throughput VASP and payment-broker nodes far from every seed are excluded. Hop distance is never a feature. We also use a public XBlock-derived Ethereum phishing benchmark \cite{yuan2020phishing}, pinned to a public repository snapshot \cite{yancykahn2023ethphishing} that identifies XBlock as its source \cite{zheng2020xblocketh}; its explicit phishing/normal labels remain unchanged. The fifteen non-label features are in- and out-degree, unique counterparties, transaction counts, value sum, mean, and standard deviation in each direction, temporal span, mean inter-transaction time, and counterparty diversity; heavy-tailed features enter as $\log(1+x)$. Table~\ref{tab:data} summarizes the scenarios, and the full labeling rules appear in the Appendix.

The benchmark spans graph sizes that differ by more than $72\times$ and labeled prevalence ranging from dense sanctions neighborhoods to only 89 positives on Russia--ETH. We treat the eight scenarios as separate deployment conditions and macro-average their repeated-run means.

\textbf{Protocol and metrics.} Every method uses the same labeled nodes and a stratified 70/15/15 split on seeds $\{0,1,2,3,42\}$; thresholds maximize validation F1. We report positive-class Precision, Recall, F1, PR-AUC, and ROC-AUC; PR-AUC complements F1 under imbalance because it does not depend on a threshold. The graph is transductive: message passing may see unlabeled topology and features, while test labels stay out of fitting, early stopping, and threshold selection. Training nodes are randomly divided into $K=3$ clients.

\textbf{Baselines.} Because the cross-VASP setting prohibits pooling raw transaction data, the comparison set consists of federated methods: FedAvg, FedProx ($\mu=0.01$), FedSage, FedGTA, FedTAD, FedTGP, FedAux, and FedGraph-VASP (Table~\ref{tab:baseline-scope}). We port their defining algorithmic components into the same GraphSAGE, split, and evaluation stack so that differences are not caused by incompatible preprocessing. All methods use class-weighted cross-entropy and validation-based threshold selection.

\begin{table}[t]
\centering
\small
\setlength{\tabcolsep}{3pt}
\caption{Federated comparison set. ``Signal'' is the distinguishing coordination mechanism retained in our common evaluation stack.}
\label{tab:baseline-scope}
\begin{tabular}{@{}llp{.52\columnwidth}@{}}
\toprule
\textbf{Method} & \textbf{Year} & \textbf{Coordination signal}\\
\midrule
FedAvg & 2017 & sample-weighted parameters; feature MLP\\
FedProx & 2020 & proximal local graph objective\\
FedSage & 2021 & subgraph GraphSAGE parameter averaging\\
FedGTA & 2023 & topology moments and personalized aggregation\\
FedTAD & 2024 & topology reliability and data-free distillation\\
FedTGP & 2024 & trainable global class prototypes\\
FedAux & 2025 & auxiliary projection and similarity weighting\\
FedGraph-VASP & 2026 & cross-silo boundary embedding exchange\\
\bottomrule
\end{tabular}
\end{table}

\textbf{Training.} Graph baselines use 100 rounds, one local epoch, a two-layer width-32 GraphSAGE backbone where the method is graph-specific, class-weighted cross-entropy, and patience-20 early stopping on validation F1. AEGIS Stage~1 copies both FedSage layers. Stage~2 uses Adam for 80 epochs, learning rate $5\cdot10^{-3}$, weight decay $5\cdot10^{-4}$, patience 20, $K=3$, $k=2$, key size 16, and residual width 16, with $\beta=0.01$ and $\gamma=0.25$. Method-specific settings and the compute environment are listed in the Appendix.

\section{Results}
\label{sec:results}

\paragraph{Overall comparison.}
AEGIS raises mean F1 from 0.751 for the strongest comparator, FedGraph-VASP, to 0.839 ($+11.8\%$ relative) and is strictly best in seven scenarios. On Russia--ETH, FedAux reaches 0.641 while AEGIS matches FedAvg at 0.604, because every AEGIS run on that graph falls below the prespecified validation-support cutoff and the guard deploys the feature-only anchor.

\paragraph{Across-scenario stability.}
Figure~\ref{fig:result-diagnostics}(a) shows that the mean margin is positive on seven graphs and negative only on Russia--ETH. Iran--TRON and Terror--TRON have the largest gains, and XBlock has a smaller gain that is positive in every run. Across the 40 seed--scenario pairs, AEGIS wins 25, ties 3, and loses 12 against the strongest comparator for each scenario. Against FedGraph-VASP, AEGIS also improves macro precision by 0.130, recall by 0.002, PR-AUC by 0.052, and ROC-AUC by 0.015 (Figure~\ref{fig:result-diagnostics}(b) and Table~\ref{tab:app-multimetric}); the PR-AUC gain shows that the improvement does not depend on the selected threshold.

\paragraph{What validation selects.}
The core receives at least 0.90 mean weight on Iran--TRON, Russia--TRON, Terror--TRON, and XBlock, whereas Iran--ETH and Terror--ETH assign 0.70 mean weight to the two anchors and Israel--TRON uses 0.65 core weight (Figure~\ref{fig:app-anchors}). On Russia--ETH the guard always activates and deploys FedAvg alone. DTEA selects its client-local track only twice, both on Iran--TRON. Deployment-level selection is therefore sparse and interpretable, and anchor use follows label support more than node count.

\begin{table}[t]
\centering
\small
\setlength{\tabcolsep}{3pt}
\caption{Strict single-component ablation over eight scenarios and repeated runs. Every row changes one component of full AEGIS; $\Delta$F1 is variant minus full.}
\label{tab:ablation}
\begin{tabular}{lrrr}
\toprule
\textbf{Variant} & \textbf{F1} & \textbf{$\Delta$F1} & \textbf{PR-AUC}\\
\midrule
\rowcolor{aegishl}\textbf{AEGIS (full)} & \textbf{.839} & $0.000$ & .890\\
w/o anchor bank & .818 & $-.021$ & .882\\
w/o low-positive guard & .837 & $-.001$ & \textbf{.891}\\
w/o DTEA & .833 & $-.006$ & .881\\
w/o Stage-1 transfer & .828 & $-.011$ & .883\\
w/o motif features & .814 & $-.025$ & .876\\
w/o residual experts & .829 & $-.010$ & .884\\
dense routing ($k=K$) & .835 & $-.004$ & .889\\
\bottomrule
\end{tabular}
\end{table}

\paragraph{Ablation.}
Every strict single-component change lowers macro F1 (Table~\ref{tab:ablation} and Figure~\ref{fig:result-diagnostics}(c)). Removing the motifs or the anchor bank costs the most, 0.025 and 0.021. Removing Stage-1 transfer costs 0.011, removing the residual experts costs 0.010, removing DTEA costs 0.006, and dense routing costs 0.004. The components help different graphs: removing motifs costs 0.164 F1 on Iran--TRON, removing the anchor bank costs 0.057 and 0.059 on Russia--ETH and Terror--ETH, and removing DTEA lowers Iran--TRON from 0.873 to 0.825, the scenario on which the local track is selected. Per-scenario results are in Table~\ref{tab:app-strict}.

\paragraph{Efficiency.}
DTEA trains two cores and raises the mean run time from 47.5 to 75.1 seconds; the two shared anchors add their own training and two forward passes. At inference only the selected DTEA track is retained, and sparse routing combines two of three residual scores while being 0.004 F1 above dense routing.

\section{Discussion}

\paragraph{Where components help.}
Motifs matter most on Iran--TRON and the anchors on the sparse Ethereum sanctions graphs. DTEA activates rarely because the standard track is already strong and its fixed margin requires clear validation evidence. The anchor bank and DTEA improve PR-AUC as well as F1, whereas the guard trades a small PR-AUC loss for F1, matching its role as a decision safeguard.

\paragraph{What should vary, and when.}
AEGIS separates two time scales. At input level, the selected experts and their weights $w_i(x)$ vary across queries within a frozen model version. At deployment level, validation selects one track, anchor vector, guard state, and threshold, which remain fixed throughout the scoring window. One-factor sweeps over $k$, $\gamma$, $\eta$, the guard cutoff, and the anchor-grid resolution show that $k=2$ is best and that only the guard cutoff has a material effect.

\paragraph{Limitations.}
The sanctions scenarios evaluate risk prioritization under weak one-hop supervision on static, transductive graphs, and clients are a balanced random partition simulated in one process. Temporal transfer, new-address induction, and consortia with observed institutional ownership are natural next steps. The released score is a case-prioritization signal for investigator review, complementary to secure aggregation, authenticated transport, and training-time privacy.

\section{Conclusion}

Collaborative blockchain screening must reconcile public transaction structure with fragmented private expertise. AEGIS combines a transferred GraphSAGE base, motif priors, residual experts, conservative dual-track adaptation, and a validation-selected anchor bank. It improves macro F1 by 11.8\% over the strongest complete federated comparator, while every strict single-component change reduces F1. The result is an auditable foundation for collaborative screening without centralizing client labels.

\bibliographystyle{ACM-Reference-Format}
\bibliography{_references}
\appendix
\begin{table*}[t]
\centering
\small
\setlength{\tabcolsep}{2.5pt}
\caption{Strict single-component test F1 per scenario. Each row changes one component of full AEGIS.}
\label{tab:app-strict}
\begin{tabular}{lrrrrrrrrr}
\toprule
\textbf{Variant} & \textbf{Iran-T} & \textbf{Russia-T} & \textbf{Terror-T} & \textbf{Israel-T} & \textbf{Iran-E} & \textbf{Russia-E} & \textbf{Terror-E} & \textbf{XBlock} & \textbf{Macro}\\
\midrule
\textbf{AEGIS (full)} & .873 & .918 & .897 & .839 & .812 & .604 & .837 & .932 & .839\\
w/o anchor bank & .869 & .900 & .897 & .832 & .786 & .547 & .778 & .932 & .818\\
w/o low-positive guard & .873 & .918 & .897 & .839 & .812 & .592 & .837 & .932 & .837\\
w/o DTEA & .825 & .918 & .897 & .839 & .812 & .604 & .837 & .932 & .833\\
w/o Stage-1 transfer & .797 & .973 & .868 & .832 & .791 & .604 & .843 & .917 & .828\\
w/o motif features & .709 & .940 & .847 & .840 & .810 & .607 & .833 & .929 & .814\\
w/o residual experts & .815 & .914 & .891 & .840 & .809 & .604 & .837 & .924 & .829\\
dense routing ($k=K$) & .849 & .918 & .895 & .840 & .804 & .604 & .832 & .935 & .835\\
\bottomrule
\end{tabular}
\end{table*}

\section{Data, Motifs, and Protocol}
\label{sec:app-protocol}

\subsection{Labels, Features, and Splits}
Sanctioned seeds come from public OFAC digital-currency designations under the IFSR, UKRAINE-EO13661/CYBER2, and SDGT programs, together with NBCTF cryptocurrency seizure orders: 592 distinct BTC, ETH, and TRON addresses, of which the experiments keep Ethereum and TRON. USDT neighborhoods are retrieved from public chain endpoints. TRON timestamps are converted from milliseconds, exact duplicate transfers are removed, invalid timestamps are dropped, negative values are clamped to zero, and values above $10^{12}$ are clipped.

Each node has one deterministic role. A listed seed is sanctioned. An unlabeled one-hop neighbor is a laundering intermediary. A high-throughput node at least two hops from every seed is a VASP account when it has at least three incoming and outgoing counterparties, lies above the 99th degree percentile (minimum 20), and has in/out value balance above 0.1. A two-hop node above the 90th degree percentile (minimum three) is a payment broker. Remaining nodes are normal users. Sanctioned and intermediary nodes are the weak positive class, normal nodes are negative, and VASP or broker nodes are excluded. On Ethereum, public tags for the 2{,}000 highest-degree addresses can override these heuristics: sanction, blocked, mixer, phishing, scam, darknet, or ransomware tags set a positive role, and exchange, bridge, or payment-processor tags set the VASP role. The public XBlock-derived benchmark lists 4{,}818 phishing and 4{,}169 normal seeds \cite{yuan2020phishing,yancykahn2023ethphishing,zheng2020xblocketh}; only those seeds are supervised.

The fifteen behavioral features are clipped and z-scored, and the seven unscaled motif scores are appended, giving a 22-dimensional input shared by AEGIS and all baselines. Every method uses the same labeled nodes and a stratified 70/15/15 split; the seeds are $\{0,1,2,3,42\}$. Training nodes are randomly partitioned across $K=3$ logical clients by a permutation seeded with the split seed plus one. The standard-track loss sums over all training nodes; the client-local DTEA loss indexes this partition.

\subsection{Motif Scores}
Let $\mathcal N^-(x)$ and $\mathcal N^+(x)$ be the in- and out-neighbors of $x$. All seven scores lie in $[0,1]$ and use only the observed ego graph.
\begin{itemize}
\item Peel chain: fraction of temporally adjacent in/out pairs, within one hour, with $0.80\leq v_{x\to w}/v_{u\to x}\leq 0.99$.
\item Fan-out: out-neighbor diversity combined with the fraction of outputs smaller than 20\% of the representative input.
\item Fan-in: the symmetric consolidation score.
\item Mixer star: $\exp[-\mathrm{CV}(v)]\min(|\mathcal N^+\cup\mathcal N^-|/50,1)$.
\item Wash cycle: fraction of out-neighbors on a length-three return path.
\item Bipartite hub: in/out value balance times log-scaled throughput.
\item Ping-pong: fraction of counterparties in both neighbor sets.
\end{itemize}

\subsection{Anchor Selection}
\begin{figure}[t]
  \centering
  \includegraphics[width=\columnwidth]{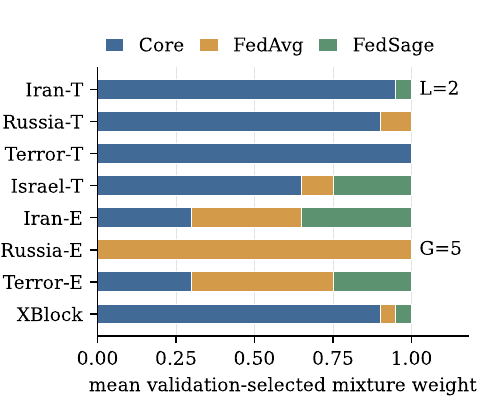}
  \Description{Average validation-selected weights of the core, feature, and graph anchors across eight scenarios.}
  \caption{Validation-selected anchor weights, averaged across runs. $L$ and $G$ count client-local DTEA selections and low-positive guard activations; omitted counts are zero.}
  \label{fig:app-anchors}
\end{figure}

Figure~\ref{fig:app-anchors} shows the validation-selected anchor weights.

\section{Settings and Compute}
\label{sec:app-compute}

Neural baselines use class-weighted cross-entropy, 100 rounds, one local epoch, a two-layer width-32 GraphSAGE backbone where the method is graph-specific, dropout 0.3 for FedSage, and validation-F1 early stopping with patience 20. FedProx ($\mu=0.01$) and FedSage ($\mu=0$) are run separately; with a weak proximal term on a balanced split, their thresholded F1 coincides on six scenarios. FedGraph-VASP is a paper-level port: balanced ownership replaces Louvain partitions, boundary endpoints are visible to both clients, and the cosine alignment weight is $\lambda=0.1$; encryption is omitted because it preserves predictions. Tables~\ref{tab:app-hyper} and~\ref{tab:app-aegis-hyper} list the final settings.

\begin{table}[t]
\centering
\small
\setlength{\tabcolsep}{2pt}
\caption{Baseline settings. Graph methods use a two-layer width-32 GraphSAGE backbone, class-weighted cross-entropy, and patience-20 early stopping on validation F1.}
\label{tab:app-hyper}
\begin{tabular}{lp{.62\columnwidth}}
\toprule
\textbf{Model} & \textbf{Settings}\\
\midrule
FedAvg & 100 rounds; 1 epoch; lr $10^{-2}$; MLP width 32\\
FedProx & FedSage path; $\mu=0.01$\\
FedSage & 100 rounds; 1 epoch; hidden 32; dropout 0.3\\
FedGTA & 5-step LP; $\alpha=0.5$; temperature 20; 10 moments\\
FedTAD & 100 pseudo-nodes; top-5; 5 server epochs\\
FedTGP & 10 prototype-server epochs; $\lambda=1$\\
FedAux & APV kernel; $\sigma=1$; similarity scale 10\\
FedGraph-VASP & boundary exchange; cosine alignment $\lambda=0.1$\\
\bottomrule
\end{tabular}
\end{table}

\begin{table}[t]
\centering
\small
\setlength{\tabcolsep}{2pt}
\caption{AEGIS settings used for every reported scenario.}
\label{tab:app-aegis-hyper}
\begin{tabular}{lp{.58\columnwidth}}
\toprule
\textbf{Component} & \textbf{Settings}\\
\midrule
Stage 1 & FedSage settings; copy both SAGE layers\\
Stage 2 & Adam; 80 epochs; lr $5\cdot10^{-3}$; weight decay $5\cdot10^{-4}$; patience 20\\
Router & $K=3$; $k=2$; key size 16; residual width 16\\
Loss & class-weighted CE; balance $\beta=0.01$; local weight $\gamma=0.25$\\
DTEA & two tracks; deploy local only if margin $\eta=0.02$\\
Anchors & FedAvg and FedSage; 0.25 simplex; guard below 20 validation positives\\
\bottomrule
\end{tabular}
\end{table}

The runs used Ubuntu 24.04, AMD EPYC 7763 CPUs, and RTX 3090 GPUs (24\,GiB), with Python 3.12.3, PyTorch 2.5.1, and PyTorch Geometric 2.7.0; one run uses one GPU.

\section{Extended Results}
\label{sec:app-results}

\begin{table}[t]
\centering
\small
\setlength{\tabcolsep}{2pt}
\caption{Metrics averaged over eight scenarios and five seeds.}
\label{tab:app-multimetric}
\begin{tabular}{lrrrrr}
\toprule
\textbf{Method} & \textbf{Prec.} & \textbf{Recall} & \textbf{F1} & \textbf{PR} & \textbf{ROC}\\
\midrule
FedAvg & .844 & .643 & .716 & .736 & .935\\
FedProx & .786 & .697 & .714 & .773 & .959\\
FedSage & .785 & .697 & .714 & .773 & .959\\
FedGTA & .777 & .750 & .739 & .811 & .973\\
FedTAD & .689 & .708 & .666 & .755 & .949\\
FedTGP & .844 & .609 & .692 & .691 & .878\\
FedAux & .810 & .682 & .731 & .781 & .950\\
FedGraph-VASP & .763 & .797 & .751 & .838 & .977\\
\textbf{AEGIS} & \textbf{.893} & \textbf{.799} & \textbf{.839} & \textbf{.890} & \textbf{.992}\\
\bottomrule
\end{tabular}
\end{table}

Table~\ref{tab:app-multimetric} reports macro metrics and Table~\ref{tab:app-strict} the per-scenario ablation.

\end{document}